# Machine Learning with Abstention for Automated Liver Disease Diagnosis

Kanza Hamid
PIEAS Biomedical Informatics Laboratory
Pakistan institute of engineering and applied sciences
Islamabad, Pakistan
kanzahamid@hotmail.com

Amina Asif
PIEAS Biomedical Informatics Laboratory
Pakistan institute of engineering and applied sciences
Islamabad, Pakistan
a.asif.shah01@gmail.com

Wajid Arshad. Abbasi
PIEAS Biomedical Informatics Laboratory
Pakistan institute of engineering and applied sciences
Islamabad, Pakistan
wajidarshad@gmail.com

Durre Sabih
Multan Institute of Nuclear Medicine and Radiotherapy
Multan, Pakistan.
dsabih@yahoo.com

Fayyaz-ul-Amir Afsar Minhas
PIEAS Biomedical Informatics Laboratory
Pakistan institute of engineering and applied sciences
Islamabad, Pakistan
fayyazafsar@gmail.com

*Abstract*— This paper presents a novel approach for detection of liver abnormalities in an automated manner using ultrasound images. For this purpose, we have implemented a machine learning model that can not only generate labels (normal and abnormal) for a given ultrasound image but it can also detect when its prediction is likely to be incorrect. The proposed model abstains from generating the label of a test example if it is not confident about its prediction. Such behavior is commonly practiced by medical doctors who, when given insufficient information or a difficult case, can chose to carry out further clinical or diagnostic tests before generating a diagnosis. However, existing machine learning models are designed in a way to always generate a label for a given example even when the confidence of their prediction is low. We have proposed a novel stochastic gradient based solver for the *learning with abstention* paradigm and use it to make a practical, state of the art method for liver disease classification. The proposed method has been benchmarked on a data set of approximately 100 patients from MINAR, Multan, Pakistan and our results show that the proposed scheme offers state of the art classification performance.

## I. INTRODUCTION

Liver diseases are a cause of major health problems and mortality especially in developing countries such as Pakistan [1]. Fatty liver disease (FLD) and heterogeneous liver texture are among the precursors of more serious liver disorders such as cirrhosis [2]. In FLD, lipid cells start accumulating in the liver whereas heterogeneous liver texture is a consequence of formation of irregular cells. The detection of these liver disorders can be difficult, especially in their initial stages [3][4]. If these conditions are not detected and treated in time, they may lead to chronic liver disease and cirrhosis which have severe health implications [5].

The most accurate method for diagnosis of such liver diseases is liver biopsy which is invasive, risky, painful and expensive [6]. Non-invasive methods for liver disease diagnosis include ultrasound (US), computed tomography (CT), elastography, etc. These methods are painless and less expensive but are also less accurate than liver biopsy [7]. The use of these diagnostic methods requires access to well-trained medical experts and diagnostic facilities. Automated diagnosis systems for liver disorders can save time and money by acting as a pre-screening service to refer only those individuals for further testing or medical advice who have a high predicted likelihood of a liver disorder.

- A number of researchers have implemented different machine learning methods to detect liver abnormalities in an automated fashion. Most of such techniques are primarily based on textural analysis of ultrasound images using statistical features followed by a machine learning classifier such as a Support Vector Machine, Random Forest or hierarchical classification, etc. [8]–[13]. Ultrasound is widely used due to its lower cost and easy availability in comparison to other more sophisticated imaging modalities such as CT or electrography. Wun et al. [8] selected statistical features such as mean, standard deviation, gray level difference, run-length percentage, entropy, etc. for ultrasound characterization and reported an agreement of 89.90% with expert classification. Badawai et al. [9] used a fuzzy logic based model for tissue characterization of liver ultrasound images. They reported specificity and sensitivity values of 92% and 96%, respectively, for fatty liver classification. Yoshida et al. [10] used multiscale texture analysis for classification of liver ultrasound images with area under a receiver operating characteristic curve (AUC ROC) of 92%. İçer et al. [11] proposed a method based on evaluation of liver enzymes with quantitative grading of fatty liver using ultrasound images. They reported AUC ROC scores of 97.5%, 95.8%, and 94.9% for normal, grade I and grade II fatty liver ultrasound images, respectively. Andrade et al. [12] applied stepwise regression as a feature selection method with k-nearest neighbor, support vector machine and artificial neural network classifiers for detection of liver steatosis using ultrasound images and reported an accuracy of 79.8%. Minhas et al. [13] proposed a wavelet transform based technique for completely automated classification of normal, heterogenous and fatty liver disorders with an accuracy of 95%. Owjimehr et al. [14] improved upon this approach by using a hierarchical classifier with an accuracy of 97.9%.

In this work, we have identified a major issue with all existing automated diagnosis methods in this domain. All existing ultrasound based liver disease diagnosis systems are designed to always generate a label for an input example even if the predicted label is highly likely to be incorrect. In contrast to existing automated techniques, a medical doctor can either choose to diagnose a patient based on current information available about the patient or, alternatively, refrain from generating any decision if the available information is not sufficient to reach a reliable diagnosis. In such cases, a doctor will typically request further diagnostic or clinical tests because the cost of a misdiagnosis can be much higher than that resulting from abstention. In the context of liver disorders, an ideal automated ultrasound based diagnosis system should follow the same pattern, i.e., it should classify an example only if it is highly confident about its prediction and should reject or abstain from classification otherwise. Such a system can function as a more effective pre-screening service in comparison to existing methods by referring only those patients for further medical examination or expensive or invasive tests such as elastography, CT or biopsy for which the classifier has abstained from classification.

With this background, we have developed an automated liver disease diagnosis system that can not only classify a given liver ultrasound as normal or abnormal but it can also refrain from classification if it is unsure about the correctness of its prediction. Our model is based on a customized implementation of the *learning with rejection* or abstention framework proposed by Cortes et al. [15]. Our experimental results on a dataset comprising of about 100 subjects collected from medical experts in Pakistan shows that the proposed learning with abstention model of automated diagnosis can very useful in practice.

The rest of the paper is organized as follows: Section II gives the details of the proposed method, Section III presents results and discussion whereas conclusions and future work are given in Section IV.

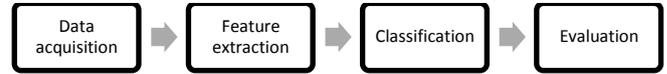

Fig. 1. Proposed methodology

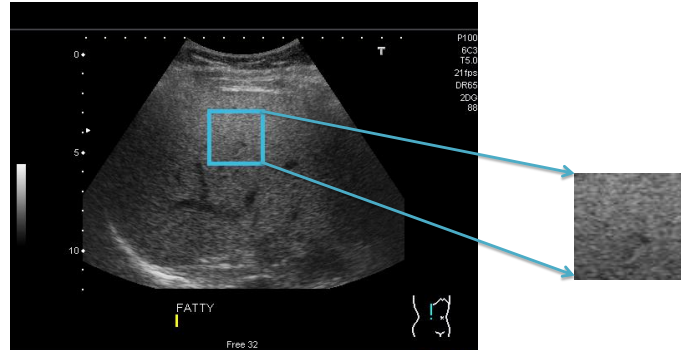

Fig. 2. Selected ROI with its ultrasound image

## II. METHODS

Our proposed methodology consists of the following steps as shown in Fig. 1: ultrasound data acquisition, region of interest (ROIs) selection and annotation by a medical expert, application of various machine learning classification techniques and performance evaluation.

### A. Data Acquisition

Our dataset consists of 99 liver ultrasound images. Among these, 43 images are of healthy individuals whereas the remaining 56 have liver abnormalities such as FLD or heterogenous liver texture. All these images were acquired at Multan Institute of Nuclear Medicine and Radiotherapy (MINAR) Multan, Pakistan, by the authors (DS) using a Toshiba Aplio 500 B-mode digital ultrasound machine. The frequency for tissue harmonic imaging was 5 MHz and a convex probe was used. The size of each acquired image is 560×450 pixels and the image was saved as a bitmap file. For these 99 images, 114 64×64 pixel region of interest (ROIs) were selected by the medical expert for annotation into normal or abnormal. All subsequent processing is done on these selected ROIs. An example of liver ultrasound image with its annotation and ROI is shown in Fig. 2. Upon publication of this article, the dataset will be made publicly available at the URL: http://faculty.pieas.edu.pk/fayyaz/software.html#LWA.

### B. Feature Extraction

All existing methods use complicated feature extraction techniques. In this work, we chose to use the normalized pixel values of the 64×64 ROIs as features. This results in a 4096-dimensional feature vector for a given example. As discussed in the results section, these simple features offer comparable or better accuracy than more sophisticated statistical features.

C. Classification

In order to test our hypothesis that learning with abstention is effective for liver disease diagnosis, we compare the performance of our implementation of *learning with abstention* with conventional classification techniques. Henceforth, we provide details of various classification methods used in this work.

*1) Nearest Neighbor (NN)*

As a baseline, nearest neighbor classifier was used to classify data into normal and abnormal classes [16]. Euclidean distance metric was used for distance calculations in the classifier.

*2) Support Vector Machine (SVM)*

A support vector machine (SVM) finds a maximum margin linear discriminant function $h(x) = w^T \phi(x) + b$ to classify the feature representation $\phi(x)$ of an example $x$ using a weight vector $w$ and a bias parameter $b$. An SVM determines the optimal values of $w$ and $b$ by using a training set $S = \{(x_i, y_i) | i = 1, 2 \dots N\}$ of examples with corresponding labels $y_i = -1$ or $y_i = +1$ for normal and abnormal cases, respectively. This is done by solving the following optimization problem:

$$min_{w,b} \frac{1}{2}\|w\|^2 + C \sum_{i=1}^{N} l_{SVM}(h, x_i, y_i)$$

Here, the first term $\|w\|^2$ is responsible for margin maximization and second term controls the number of misclassification over training data by using hinge loss function $l_{SVM}(h, x_i, y_i) = max\{0, 1 - y_i h(x_i)\}$. The hinge loss function penalizes misclassifications and margin violations. The hyper-parameter $C$ is the weighting factor between these two terms and is chosen through cross-validation [17].

*3) Learning with Abstention (LWA)*

Conventional classifiers are designed to always produce a label given an example which can either be correct or incorrect. As discussed earlier, it would be more practical if a classifier can abstain from generating a label when it is not confident about its decision instead of a producing a misclassification. In this work, we have implemented a classifier that can refrain from generating labels for such test examples. The idea of learning with abstention was proposed by Cortes et al. [15]. Such a classifier can generate three different types of labels in our case: normal ($-1$), abnormal ($+1$) or Reject ($R$) which corresponds to an abstention from classification. We follow the same principle for construction of the LWA classifier as in Cortes et al [15]. However, unlike their approach, we have solved the optimization problem of the LWA classifier using a stochastic gradient based solver [18].

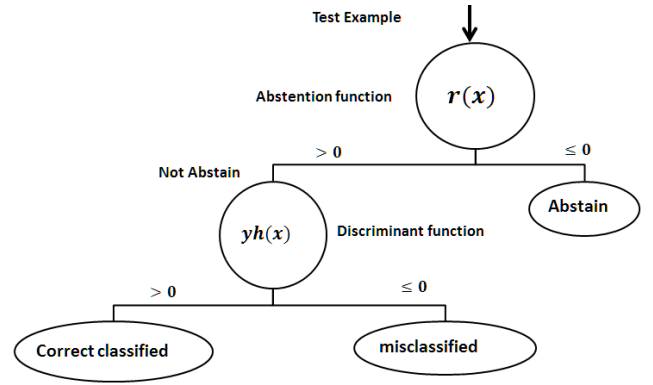

Fig. 3. Systematic diagram of proposed method

As discussed in the work by Cortes et al. [15], LWA requires two decision functions: a discriminant function $h(x)$ which is the same as in a standard SVM and an abstention function $r(x) = u^T \phi(x) + b'$ that uses different weight and bias parameters. The objective of LWA is to simultaneously learn both these functions in a way that the rejection function produces a positive score $r(x) > 0$ only if the discriminant function is expected to correctly classify the given example. If $r(x) < 0$, the classifier is not confident about the correctness of the label generated by its discriminant function and the example is rejected (abstention). A systematic representation of this concept is shown in Fig. 3.

Similar to a conventional SVM, a large margin LWA classifier can be developed through the principle of structural risk minimization [16] by simply using a loss function that takes abstentions into account. For implementation of the LWA classifier, we use the loss function and its convex over-approximation given by Cortes et al. [15] which works as follows:

i. *Correct classification without rejection*: In the scenario in which an example is not rejected ($r(x) > 0$) and is classified correctly ($yh(x) > 0$), no loss is incurred.

ii. *Misclassification without rejection*: An example that is not rejected ($r(x) > 0$) but is misclassified ($yh(x) \leq 0$), incurs a loss of 1.0.

iii. *Abstention:* The abstention or rejection of an example ($r(x) < 0$) incurs a loss of $c \in (0, 0.5)$. The hyper-parameter $c$ is set by the user and it controls the cost and, consequently, the number of rejections. A small $c$ will produce more rejections and vice versa.

The loss function can be written mathematically formulated as follows:

$$l(h, r, x, y) = \mathbb{I}(yh(x) \leq 0)\mathbb{I}(r(x) > 0) + c\mathbb{I}(r(x) \leq 0) \quad (1)$$

Here $\mathbb{I}(\cdot)$ is the indicator function whose value is 1.0 if its argument is true and 0.0 otherwise. This loss function is non-linear, non-convex and difficult to optimize. Its convex over-approximation can be written as [15]:

$$l_{LWA}(h, r, x, y) = \max\left(0, 1 + \frac{1}{2}\big(r(x) - yh(x)\big), c\big(1 - \beta r(x)\big)\right) \quad (2)$$

Here, $\beta = \frac{1}{1-2c}$. Notice that this loss function will always penalize abstentions ($r(x) < 0$) and misclassifications

($yh(x) < 0$). For a more detailed description of the loss function, the interested reader is referred to [15].

Following the principle of structural risk minimization and using the above loss function, the LWA learning problem can be expressed as the following optimization problem:

$$\min_{w,u,b,b'} J(w, u, b, b') = \frac{\lambda}{2}\|w\|^2 + \frac{\lambda'}{2}\|u\|^2 + \sum_{i=1}^{N} l_{LWA}(h, r, x_i, y_i) \quad (3)$$

Here, the first two terms control the margin for the discriminant and rejection functions using hyper-parameters $\lambda$ and $\lambda'$ whereas the second term is responsible for loss-minimization. The solution to this optimization problem will result in optimal values of weights for both decision functions so that both misclassifications and abstentions are minimized. We have developed a stochastic gradient solver for the LWA optimization problem in equation (3). The proposed algorithm is inspired from the Pegasos solver for conventional support vector machines proposed by Shalev-Shwartz et al. [18]. It offers an easier and more scalable alternative to quadratic programming or sequential minimal optimization methods typically used in SVMs. The proposed method is based on step-wise iterative updates to weight parameters in a direction opposite to the sub-gradients of the objective function using a single randomly chosen training example. The weight update equations at iteration $t$ can be written as follows:

$$w_{t+1} = w_t - \eta \nabla_w \quad (4)$$
$$u_{t+1} = u_t - \eta' \nabla_u \quad (5)$$

Here, $\eta = \frac{1}{\lambda t}$ and $\eta' = \frac{1}{\lambda' t}$ are the step-sizes for the gradients $\nabla_w = \frac{\partial J}{\partial w}$ and $\nabla_u = \frac{\partial J}{\partial u}$, respectively. For a randomly chosen training example $x$ with label $y$, the sub-gradients of the objective function can be computed by taking the derivative of the objective function with respect to the weight parameters. Consequently, the sub-gradients can be written as follows:

$$\nabla_w = \begin{cases} \lambda w - \frac{1}{2} y\phi(x) & \text{if } 1 + \frac{1}{2}(r(x) - yh(x)) > \max\left(0, c(1 - \beta r(x))\right) \\ \lambda w & \text{else} \end{cases}$$

$$\nabla_u = \begin{cases} \lambda' u + \frac{1}{2}\phi(x) & \text{if } 1 + \frac{1}{2}(r(x) - yh(x)) > \max\left(0, c(1 - \beta r(x))\right) \\ \lambda' u - c\beta\phi(x) & \text{if } c(1 - \beta r(x)) > \max\left(0, 1 + \frac{1}{2}(r(x) - yh(x))\right) \\ \lambda' u & \text{else} \end{cases}$$

Substituting the above sub-gradient calculations into the weight update equations leads us to the complete algorithm for learning with abstention which is given in Fig. 4. It is important to note that the bias term has been omitted for clarity and it is trivial to obtain bias update equations. As discussed earlier, the proposed algorithm operates by selecting a training example from the training data uniformly at random and calculating the sub-gradient of the objective function and performing weight updates in the direction opposite to the sub-gradient. The hyper-parameters $\lambda, \lambda'$ and $c$ are selected through cross-validation. Once the optimal weight vectors have been obtained, the classifier can generate labels for a given test example: if $r(x) < 0$, the example is rejected as the classifier is not confident about its prediction, otherwise, the decision function $h(x)$ is used to determine the class (normal or abnormal) for the given example. This algorithm has been implemented in Python 2.7. Upon publication of this article, code will be made publicly available at the URL: http://faculty.pieas.edu.pk/fayyaz/software.html#LWA.

D. Evaluation

For evaluation of performance of all classifiers used in this work and their comparison with existing methods, we have used leave one out cross validation. Leave one out cross validation [19] is the method of choice for evaluating machine learning problems with small data sets. In this approach, a single example is held out as a test case while the model is trained on all other examples and this process is repeated for all examples. As performance metrics, we have used accuracy and area under the receiver operating characteristic curve (AUC-ROC) as well as the number of misclassifications and abstentions. AUC-ROC is obtained by plotting the specificity of the classifier at different decision thresholds vs. its sensitivity. The higher the value of AUC-ROC, the better the classifier [19].

```
Learning with Abstention Using Stochastic Gradients
INPUT: Training set S = {(x_i, y_i)|i = 1, 2 … N}
HYPER-PARAMETERS:
  Regularization parameter for w: λ > 0
  Regularization parameter for u: λ' > 0
  Abstention Penalty: c ∈ (0,0.5)
  Number of iterations: T > 0
INITIALIZE: w_1 = 0 , u_1 = 0 , β = 1/(1 − 2c)
For t = 1, 2, …, T
  Choose example (x, y) ∈ S uniformly at random
  Calculate h(x) = w_t φ(x)
  Calculate r(x) = u_t φ(x)
  If (1 + ½(r(x) − yh(x))) > max(0, c(1 − βr(x))) then:
    w_{t+1} ← (1 − 1/t)w_t + 1/(2λt) yφ(x)
    u_{t+1} ← (1 − 1/t)u_t − 1/(2λ't) φ(x)
  ElseIf c(1 − βr(x)) > max(0, 1 + ½(r(x) − yh(x))) then:
    w_{t+1} ← (1 − 1/t)w_t
    u_{t+1} ← (1 − 1/t)u_t + cβφ(x)
  Else:
    w_{t+1} ← (1 − 1/t)w_t
    u_{t+1} ← (1 − 1/t)u_t
OUTPUT:  w = w_{T+1} , u = u_{T+1}

Classification with Abstention Using Stochastic Gradients
INPUT: Test example (x, y)
Calculate h(x) = wφ(x)
Calculate r(x) = uφ(x)
If r(x) < 0:
  Output "Reject"
Else:
  Output h(x)
```

Fig. 4. Pseudo code of proposed classifier

III. RESULTS AND DISCUSSION

A. Comparison of LWA with conventional classification

Table 1 and Figure 5 show the results of different classification techniques in terms of AUC-ROC and accuracy.

The nearest neightbor classifier gives AUC-ROC and accuracy of 57% and 82%, respectively, and forms the baseline for comparison with other methods. The conventional support vector machine performs significantly better with an AUC-ROC of 84% and comparable accuracy (~80%).

In order to compare the performance of the proposed LWA classifier, we refer to Figure 5 which plots both the AUC-ROC and the fraction of abstentions vs. the absetention cost parameter $c$. The AUC-ROC of NN and SVM are also plotted as reference. It can be noticed that the AUC-ROC of LWA is always better than or comparable to the conventional SVM. As expected, the increase in abstention penalty decreases the fraction of absetention: for low value of $c = 0.1$, the LWA classifier rejects all examples whereas for high value of $c = 0.5$, no abstentions take place. Furthermore,. As expected, when the fraction of absetention drops to zero for large values of $c$, the performance of LWA becomes comparable to a conventional SVM. However, for $c = 0.45$, the fraction of abstention is equal to 7% with an AUC-ROC of 99%. This shows that the LWA classifier achieves near perfect classification accuracy if it is permitted to abstain from producing labels for 7 test examples. Instead of generating wrong labels for these 7 test examples as done by the conventional SVM and nearest neighbor classifiers, LWA has automatiically detected that its confidence for correctly predicting these examples is low and it has abstained from these misclassifications. This shows the effectiveness of the proposed approach in comparison to conventional classification techniques. The python implementation of the LWA classifier runs in under 5-6 minutes on a laptop with an Intel core i5-3317U 1.70 GHz processor and 4 GB RAM.

B. Re-Evaluation of rejected examples by medical expert

The 7 test cases for which the LWA classifier abstained from generating labels were given to an experienced radiologist (DS) for re-evaluation. The radiologist was not provided the original labels for these cases and was asked to diagnose these cases. It is interesting to notice that, for 3 out of these 7 cases, the radiologist generated labels different from the original labels. These cases are shown in Figure 6. This shows that the abstentions produced by the proposed LWA method were indeed difficult to classify even for trained medical experts. These cases can referred to further testing through elastography, CT or biopsy. These results clearly indicate the effectiveness of the proposed approach.

C. Comparison with existing methods

We compare the performance of our method to the one proposed by Owjimehr et al. [14] because they have used the same dataset and evaluation protocol and offer state of the art accuracy. These results are shown in Table 1. It is important to note that the approach by Owjimehr et al. [14] uses conventional classification techniques with a sophisticated feature extraction step. Table 2 shows the proposed LWA model gives better accuracy than the previous state of the art using normalized raw pixel values as features for normal vs. abnormal classification.

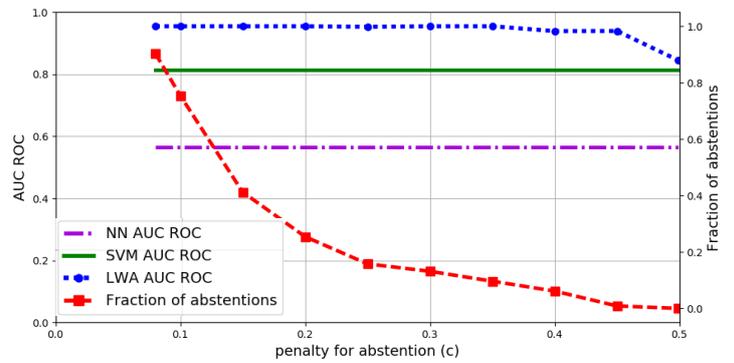

Fig. 5. Results of proposed method

**Table 1**: Comparison of proposed model with Owjimehr et al. [14]

| Methods | Owjimehr et al. [14] | | Proposed method | | |
|---|---|---|---|---|---|
| Classifier | NN | SVM | NN | SVM | LWA |
| # Misclassifications | 6 | **2** | 21 | 23 | **2** |
| # Abstentions | N/A | | N/A | | 7 |
| AUC ROC | 92 | 97 | 57 | 84 | **99** |
| Accuracy (%) | 93 | 97 | 82 | 80 | **98** |

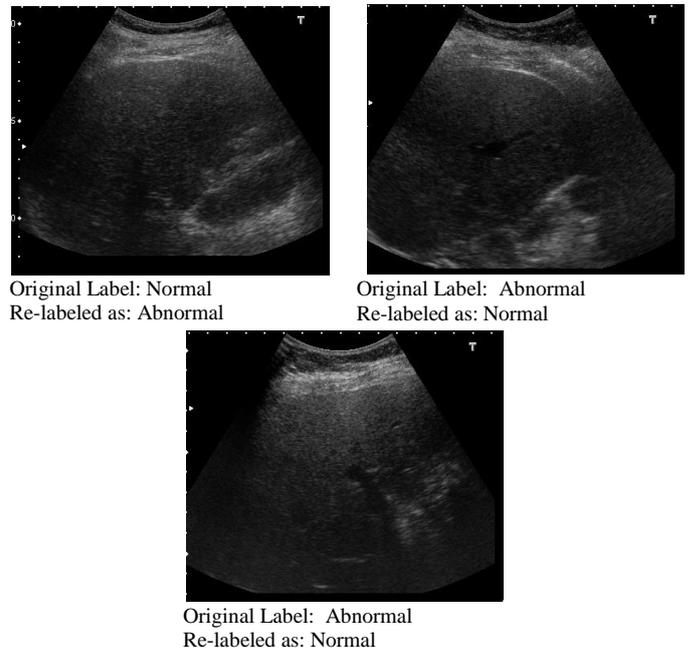

Original Label: Normal
Re-labeled as: Abnormal

Original Label: Abnormal
Re-labeled as: Normal

Original Label: Abnormal
Re-labeled as: Normal

Fig. 6. Results of re-labeling by a medical expert of the LWA-rejected images

IV. CONCLUSIONS AND FUTURE WORK

In this work, we have developed a novel approach for classification of liver ultrasound images into normal and abnormal cases. The novelty of our approach lies in using a *learning with abstention* model for classification. Our proposed method is able to automatically identify cases for which it does not have high enough confidence of generating accurate predictions. Thus, the model can be thought of an artificial intelligence (AI) system that *knows what it doesn't know*. Our results clearly show that the proposed system is very useful in a practical setting and can help both patients and medical doctors by saving them time, money and the inconvenience of undergoing painful or expensive tests. We have also proposed a novel stochastic gradient based solver for the LWA framework. The proposed scheme can be applied in other domains as well. In future, we aim to extend this method to multi-class classification and evaluate our performance on a large independent test set with elastography data. We also plan to build a publicly accessible webserver implementation of our method.

**Author contributions**: **KH** – Implementation, testing and manuscript writing, **AA** – Mathematical formulation and prototype development, **WA** – Data collection server development, **DS** – Medical data collection, annotation and supervision, **FuAAM** – Original idea, supervision and manuscript writing.

**Acknowledgements:** KH and AA are supported by MS and PhD scholarships, respectively, from the Information Technology and Telecom Endowment Fund, PIEAS. WA is supported by a PhD grant from the higher education commision under the 5000 indigenous Ph.D. scholars scheme.